# Bio-inspired Night image Enhancement based on Contrast Enhancement and Denoising

Xinyi Bai[1,2], Steffi Agino Priyanka[2], Hsiao-Jung Tung[2], Yuan-Kai Wang[2]

[1] University of Electronic Science and Technology of china, Chengdu, China.
`babsbxy@outlook.com`
[2] Fu Jen Catholic University, Taiwan.
`steffi@islab.tw, ykwang@fju.edu.tw`

**Abstract.** Due to the low accuracy of object detection and recognition in many intelligent surveillance systems at nighttime, the quality of night images is crucial. Compared with the corresponding daytime image, nighttime image is characterized as low brightness, low contrast and high noise. In this paper, a bio-inspired image enhancement algorithm is proposed to convert a low illuminance image to a brighter and clear one. Different from existing bio-inspired algorithm, the proposed method doesn't use any training sequences, we depend on a novel chain of contrast enhancement and denoising algorithms without using any forms of recursive functions. Our method can largely improve the brightness and contrast of night images, besides, suppress noise. Then we implement on real experiment, and simulation experiment to test our algorithms. Both results show the advantages of proposed algorithm over contrast pair, Meylan and Retinex.

## 1  Introduction

Surveillance system has been widely used to do security work. While the accuracy of these algorithms is largely based on the quality of input videos. Due to the lack of exposure, nighttime images are characterized as low brightness, low contrast, and high noise. This paper proposed a novel bio-inspired chain method to deal with them.

Human eyes can figure out information from different scales and exposure levels quickly and then adapt their pupils to the environmental brightness in order to take corresponding strategies for further detection automatically, because many receptive fields exist in their retinas, which helps to express the overall characteristics from positions under various exposure level. And these complete receptive fields are crucial for attention selection, which makes human eyes only focus on the object and ignore unrelated background and noise. So the first contribution in this paper is to divide nighttime images into three levels as three pseudo receptive fields: low light level (LLL), very low light level (VLLL), and high dynamic range (HDR). According to the three pseudo receptive fields, we simulate three levels of images and enhance the contrast and brightness using different parameters. The noise in nighttime images mainly includes Poisson noise and false color noise, so we use bilateral filter as fundamental pseudo attention selection to remove noise in R, G, B channels respectively.

Visual information propagates from the retina issued through the lateral geniculate nucleus finally reaches the primary visual cortex (V1), then signal processing is divided into two pathways, ventral pathway and dorsal pathway. And ventral pathway passes through V2 region to V4 region, finally arrive at inferior temporal cortex to deal with color and shape information, while dorsal pathway passes through V2 region to V3 region and middle temporal cortex, finally reaches posterior parietal cortex to deal with spatial information. So the second contribution in this paper is to divide our contrast enhancement algorithms into two simultaneously pathways, one focuses on the color, the other focuses on the brightness. Besides, human visual system is a hierarchical structure, the latter layer accepts the information processed from the former layer, and performs more advanced operation it, so every pathway of our method is a chain of algorithms. Another contribution in this paper is to count the pros and cons that do contrast enhancement before denoising and denoising before contrast enhancement. Therefore, in this paper, we also analyze and discuss the sequences of contrast enhancement algorithm and denoising algorithm.

In recent years, researches about contrast enhancement mainly include self-enhancement and context-based fusion. For context-based fusion, it aims to fuse multiple images into one image, and it is commonly applied to high dynamic range images such as denight [1], which fuses the night image with corresponding day images under abundant exposure. But this method is applicable to day and night scenes without objects. Because if there are objects in the day image, through denight contrast enhancement, the object will disappear; if there are object in the night image, the enhanced image will have ghosts. Nevertheless, when color light sources exist in the image, the color light sources will lead to color shift [1]. For self-enhancement, the global method such as contrast pair [3] is to obtain the ratio of the target point pixel value and neighboring pixel values, then calculates the mapping function based on these ratios. If the brightness is the same, the change of these pixel values in different region remains a constant, but in local methods such as Retinex and dehaze, the change of the pixel values is related to the neighboring pixel values [4-7]. Moreover, denoising algorithm mainly includes spatial method and spatio-temporal method. Common spatial method contains traditional Gaussian filter, median filter, bilateral filter and non-local mean filter [8, 9]. While spatiotemporal methods in [2, 9, 10] use the filter from the radio based method, which extends the two dimensional filter into three dimensional filter, and takes the combination of the initial video with the enhanced video as the reference of the denoising algorithm.

Main contributions of the bio-inspired image enhancement algorithm are to analogy the ventral and dorsal parallel pathways in visual signaling, and three receptive fields in retina, besides, use chain of contrast enhancement algorithms to analogy a hierarchical structure in human visual system.

## 2    Contrast enhancement and denoising

We establish a night image enhancement framework based on contrast enhancement and denoising. Our proposed method aims to convert image $I_N(x,y)$ which has low brightness, low contrast and high noise to a brighter and clearer image $I_{NIE}(x,y)$.

In the chains of contrast enhancement algorithm, we simultaneously implement it with two parallel pathways like the ventral and dorsal pathways. On one pathway, we separate the luminance channel $I_N^L$ from the night image, then improve the global contrast using tone mapping, next we do RBAF on the logarithmic domain to correct the brightness and gain the enhanced luminance channel $I_{CE}^L$, finally with the method of Histogram smoothing, the over strengthened pixels at the peak of Histogram will be solved. On the other pathway, we focus on keeping the consistence of the brightness in color channels, first we obtain a new brightness channel $I_N'$ in the global image, then we separate another two color channels $I_N^{C_1}$ and $I_N^{C_2}$ in the logarithmic domain through PCA conversion, next by multiplying them with a weight α, the color saturation is improved. Finally, we convert the enhanced brightness channel $I_{CE}^L$ and two color channel $I_N^{C_1}$, $I_N^{C_2}$ back to RGB color space. In denoising algorithm, considering the main noise in the night image is false color noise, we use bilateral filter to remove the noise in three channels respectively from image $I_{CE}$. Finally, in the enhanced images, the strength of edges will be maintained, the contrast and brightness can be largely improved, and the noise will be removed. The flow chart of our algorithm is shown in **Fig. 1**.

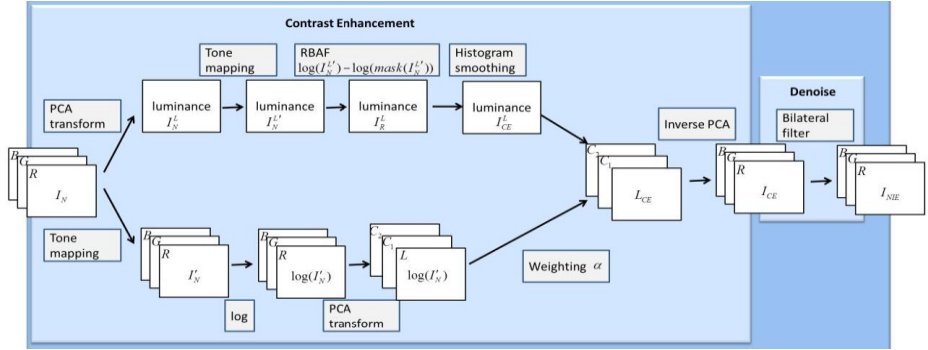

**Fig. 1.** Night image enhancement flow chart.

First we convert night image $I_N$ from RGB to $L, C_1, C_2$. Then we use the average brightness $\overline{I_N^L}$ to adjust the slope in tone mapping, where $\overline{I_N^L}$ ranges in [0, 1]. From (1), (2), $\frac{1}{\gamma}$ is the average brightness ranging in [ε, 1], and the adjusting coefficient $\lambda = \frac{1}{6}$, $\varepsilon = \frac{2}{3}$. Then in (3), p is the value of the pixel, N is the total pixels of the image. After tone mapping, RBAF is adapted for local adjustment. In (4), (5), (6), the coefficient $\beta(x,y)$ is to maintain the consistence of bright areas and dark areas. Where $\beta(x,y)$ is determined by $I_N^{L'}(x,y)$, and $\beta(x,y)$ is close to 0 in the bright area, while in the dark area, $\beta(x,y)$ is close to 1.

$$I_N^{L'} = (I_N^L)^{1/\gamma} \tag{1}$$

$$\frac{1}{\gamma} = \min(1, \Gamma \overline{I_N^L} + \varepsilon) \tag{2}$$

$$\overline{I_N^L} = \frac{\sum_{p \in I_N^L} \log(I_N^L(P))}{N} \tag{3}$$

$$R(x,y) = \log\left(I_N^{L'}(x,y)\right) - \log\left(mask\left(I_N^{L'}(x,y)\right)\right) \tag{4}$$

$$R(x,y) = \log\left(I_N^{L'}(x,y)\right) - \beta(x,y) \cdot \log\left(mask\left(I_N^{L'}(x,y)\right)\right) \tag{5}$$

$$\beta(x,y) = 1 - \frac{1}{1+e^{-7\left(I_N^{L'}(x,y)-0.5\right)}} \tag{6}$$

To deal with the halo artifacts in Retinex, in (7), we replace Gaussian filter to adaptive filter, let the contour of the object in accordance with the image itself. And change the $\sigma_{\theta,r}$ in Gaussian filter to kinetic energy as $mask(I_N^{L'}(x,y))$, where $\sigma_{\theta,r}$ is determined by whether the edge of the $mask(I_N^{L'}(x,y))$ has a high contrast. Namely, we scan the edge in eight directions from the mask, when it matches the edge precisely, $\sigma_{\theta,r}^2 = \sigma_1$; when it failed to match the edge and fell to the smooth area, $\sigma_{\theta,r}^2 = \sigma_0$, where $r$ is the distance from the point to the center of the mask, $\theta$ is the angle of Retinex based adaptive filter, when $0 < \sigma_1 \leq \frac{1}{2}\sigma_0$, the halo artifacts can be largely reduced as (8). If the value of the reflection image from (5) is less than 0, histogram stretching is required to do normalization. As is shown in (9), min and max mean the maximum and minimum values selected by R(x, y).

$$mask(I_N^{L'}(x,y)) = \frac{\sum_{\theta=0}^{360}\sum_{r=0}^{rmax} I_N^{L'}(x+r\cos(\theta), y+r\sin(\theta)) \cdot e^{\frac{r^2}{\sigma_{\theta,r}^2}}}{\sum_{\theta=0}^{360}\sum_{r=0}^{rmax} e^{\frac{r^2}{\sigma_{\theta,r}^2}}} \tag{7}$$

$$\sigma_{\theta,r}^2 = \begin{cases} \sigma_0, no\ high-contrast\ edge\ was\ crossed\ along\ \theta \\ \sigma_1, a\ high-contrast\ edge\ was\ crossed\ along\ \theta \end{cases} \tag{8}$$

$$I_R^L(x,y) = \frac{R(x,y) - min}{max - min} \tag{9}$$

Since the normalization will reduce the contrast of some areas, we use histogram smoothing to improve it. The histogram of the input image $h_i$ is changed into $h$. And we come to its cumulative histogram $H(i)$, then take $H(i)$ as the mapping function, as (10), (11). Nest, in denoising, as (12), the intensity value at each pixel in an image is replaced by a weighted average of intensity values from nearby pixels. These weights are in Gaussian distribution.

$$I_{CE}^L(x,y) = H(i)I_R^L(x,y) \tag{10}$$

$$h = \left((1+\lambda)I + \gamma D^T D\right)^{-1}(h_i + \lambda u) \tag{11}$$

$$I_{CE}(x,y) = W^{-1}\left[I_{CE}^L(x,y) + \alpha\left(I_N^{C1'}(x,y) + I_N^{C2'}(x,y)\right)\right] \tag{12}$$

Finally, the noise in the enhanced image $I_{CE}$ is removed. In our output image $I_{NE}(x,y)$, the contrast and brightness are largely improved, besides, the noise is removed, and the sharp edge remains clear.

## 3   Simulation experiment and analysis

This paper selects experiment images from TID2008 datasets [11]. We divide night images into three levels like three pseudo receptive fields, low light level (LLL), whose contrast may be well enough in the original image, but the low contrast in night time makes us difficult to recognize; very low light level (VLLL), its dark pixels in the original image do not clear enough, then global image become so dark, it is impossible to recognize; and high dynamic range (HDR), where the existing ambient light source in the image makes partial areas extremely bright or dark. In order to simulate the noise, we bring in colorful Poisson noise. The process of simulating three levels of images is divided into two steps, first decrease the brightness of the global image to $d$ to simulate a certain proportion of distortion t in dark area. Then proportionally reduce the kinetic energy according to the pixel values. In (13), $f(x)$ is the simulation image, $\alpha$ is the reduced weight of the kinetic energy, $d$ is the reduced brightness, which is deduced from (14), calculate the cumulative distribution of the distortion region $t$ in dark areas, where $0 < \alpha < 1$, $0 \leq t \leq 1$.

$$f(x) = \alpha\big(max(x-d, 0)\big) \tag{13}$$

$$d = min\{i | cdf(i) \geq t\} \tag{14}$$

The distortion t of the LLL is 0.03, α is 0.7; while the distortion t of VLLL is 0.03, α is 0.3. In the histogram of the simulated images, where the average of brightness in VLLL is 0.077115, the standard deviation is 0.055803. We also simulate the HDR image, from (15), (16), (17), (18), where $t_{low}$ and $t_{high}$ are both 0.05.

$$f(x) = min\big(\alpha\big(max(x - d_{low}, 0)\big), 1\big) \tag{15}$$

$$d_{low} = min\{i | cdf(i) \geq t_{low}\} \tag{16}$$

$$d_{high} = max\{i | cdf(i) \leq 1 - t_{high}\} \tag{17}$$

$$\alpha = \big(d_{high} - d_{low}\big)^{-1} \tag{18}$$

We use bilateral filter to denoise and compare our contrast enhancement algorithm under VLLL with contrast pair, Retinex, and Melan. In contrast pair, the edge threshold is 10; in Meylan, $\sigma_0 = 16$, $\sigma_1 = 5$, $\alpha = 1.6$; in our method, $\sigma_0 = 16$, $\sigma_1 = 5$, $\alpha = 1.6$, $\lambda = 1$, $\gamma = 1$.

Few researchers test implement orders of the two algorithms, we also analyze the sequences based on LLL-Poisson simulation: denoising after contrast enhancement (Bilateral-CE), and contrast enhancement after denoising (CE-Bilateral). Results are shown in **Fig. 2**, it is clearer on the edge of the hat and shade region in CE-Bilateral. Moreover, the contrast of the shade from contrast pair is lower than our method; the dark region in Retinex is in bluish hue compared with our method; the contrast is higher in the bright area from our method than Meylan. So our method not only improves the brightness, but also provides good contrast in dark areas. And for sequences, our method performs better under CE-bilateral than under bilateral-CE.

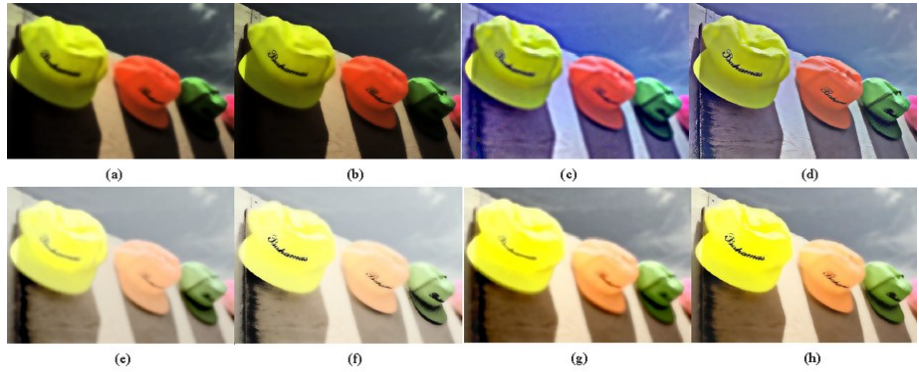

**Fig. 2.** Comparison of CE-Bilateral and Bilateral-CE based on LLL-Poisson, (a)(c)(e)(g) Bilateral-CE, (b)(d)(f)(h) CE-Bilateral, (a)(b) contrast pair, (c)(d) Retinex, (e)(f) Meylan, (g)(h)proposed.

## 4    Real night experiments and analysis

We use SSIM, Luminance and VCM [12] to compare our algorithm with contrast pair, Meylan and Retinex in real night experiments.

As is shown in **Fig. 3**, **Fig. 4,** we use digital single lens reflex camera (DSLR) to take VLLL image. Experimental results show that the four methods improve the brightness and contrast. However, contrast pair has the least contrast and brightness among them; Retinex performs better in dark areas but less in bright areas compared with ours, besides, it causes severe noise, the results of our method and Meylan are both great, but our method improves the contrast of dark areas better than Meylan. Moreover, our method enjoys the highest VCM.

Next, we implement on a second experiment to compare our method with denight and contrast pair, where the edge threshold is 10 in contrast pair; in our method, $\sigma_0 = 10$, $\sigma_1 = 3$, $a = 1.6$, $\alpha = 1$, $\lambda = 1$, $\gamma = 1$; in bilateral filter, $w = 15w$, $\sigma_d = 0.5$, $\sigma_r = 3$; in Gaussian filter, $\gamma = 5w = 9$, $\sigma_d = 0.3w$, $\sigma_r = 0.15$. As is shown in **Fig. 5**, **Fig. 6**.

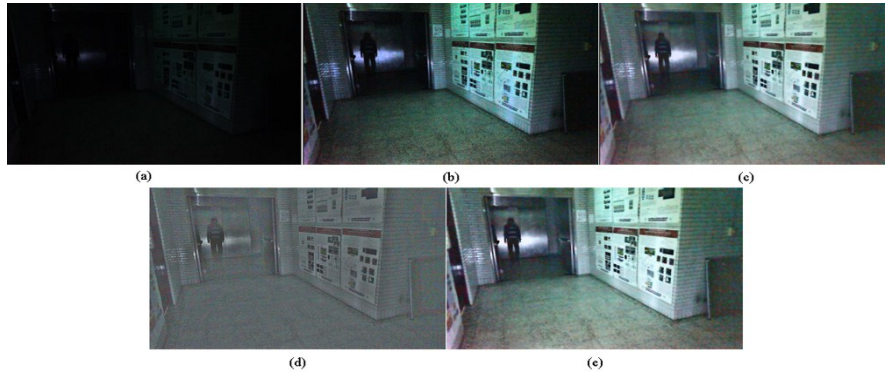

**Fig. 3.** VLLL night image enhancement I, (a) source, (b) contrast pair, (c) Meylan, (d) Retinex, (e) proposed.

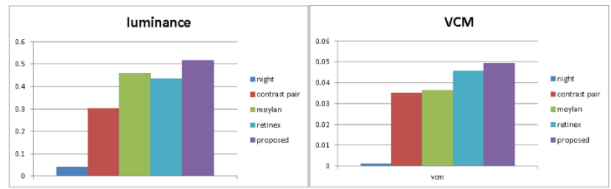

**Fig. 4.** Quantitative analysis on four algorithms.

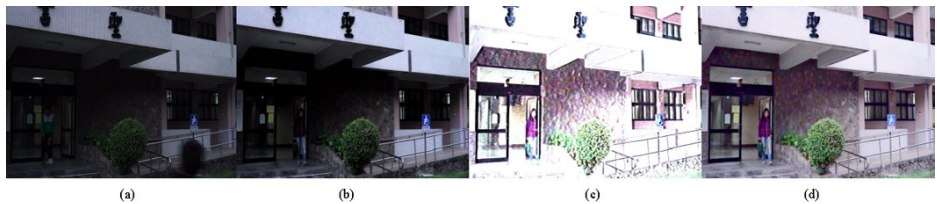

**Fig. 5.** VLLL night image enhancement II, (a) source night image, (b) contrast pair, (c) denight, (d) proposed.

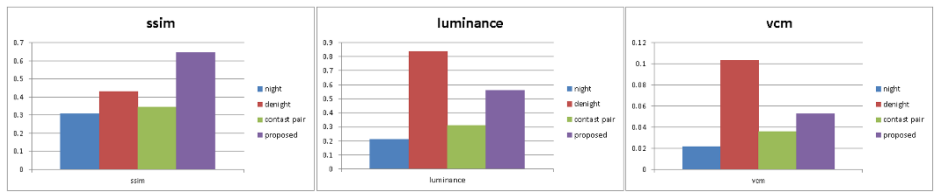

**Fig. 6.** Quantitative analysis on three algorithms.

In the final analysis, the three algorithms all improve the SSIM, brightness and contrast of the image, our method is better than contrast pair in VCM, and performs best in SSIM among them. Denight results in severe ghosts and noise. Our method largely improves the contrast and brightness of night images and nearly has no noise.

## 5      Conclusion

We use chains of image processing algorithms to construct a bio-inspired night image enhancement algorithm including ventral and dorsal parallel pathways and three pseudo receptive fields, which aims to solve the problem of low brightness, low contrast and high noise in night images. First we implement contrast enhancement algorithm dealing with low brightness and low contrast. Then we adopt bilateral filter to remove the noise in the enhanced image. Next we simulate LLL, VLLL, HDR night images and confirm that doing contrast enhancement before denoising is the best. Next, we compare proposed method with various algorithms on real night experiment. Finally, our proposed method provides the best results, which puts out image with higher brightness, higher contrast, less noise, and objects are in no ghosting condition.